\documentclass[10pt,twocolumn,letterpaper]{article}

\usepackage[pagenumbers]{cvpr} %
\usepackage{tabularx}

\usepackage{pifont}
\newcommand{\cmark}{\ding{51}}%
\newcommand{\xmark}{\ding{55}}%
\renewcommand{\vec}[1]{\mathbf{#1}}
\newcommand{\mat}[1]{\mathtt{#1}}

\newcommand{\myref}[1]{(\ref{#1})}
\newcommand{\myeqref}[1]{Eq.~\myref{#1}}
\newcommand{\myfigref}[1]{Fig.~\ref{#1}}
\newcommand{\mytabref}[1]{Tab.~\ref{#1}}

\newcommand\customparagraph[1]{\vspace{0.4em}\noindent\textbf{#1.}}

\newcommand{\mycaption}[2]{\caption[#1]{\textbf{#1} #2}}

\definecolor{cvprblue}{rgb}{0.21,0.49,0.74}
\definecolor{ourGreen}{RGB}{0,200, 0}
\definecolor{tabRed}{RGB}{170,80, 0}
\definecolor{tabGreen}{RGB}{0, 130, 0}
\definecolor{tabBlue}{RGB}{0, 0, 170}
\usepackage[pagebackref,breaklinks,colorlinks,allcolors=cvprblue]{hyperref}
\usepackage{multirow}

\def\paperID{3930} %
\def\confName{CVPR}
\def\confYear{2025}

\title{R-SCoRe: Revisiting Scene Coordinate Regression for Robust Large-Scale Visual Localization}

\author{
    Xudong Jiang$^1$
    \quad
    Fangjinhua Wang$^1$\footnotemark[1]
        \quad
        Silvano Galliani$^2$
        \quad
        Christoph Vogel$^2$
        \quad
        Marc Pollefeys$^{1,2}$\\
        $^1$Department of Computer Science, ETH Zurich\\
        $^2$Microsoft Spatial AI Lab, Zurich}

\begin{document}
\twocolumn[{%
\renewcommand\twocolumn[1][]{#1}%
\maketitle

\begin{center}
    \centering
    \vspace{-0.67cm}
    \captionsetup{type=figure}
    \includegraphics[width=0.57\linewidth]{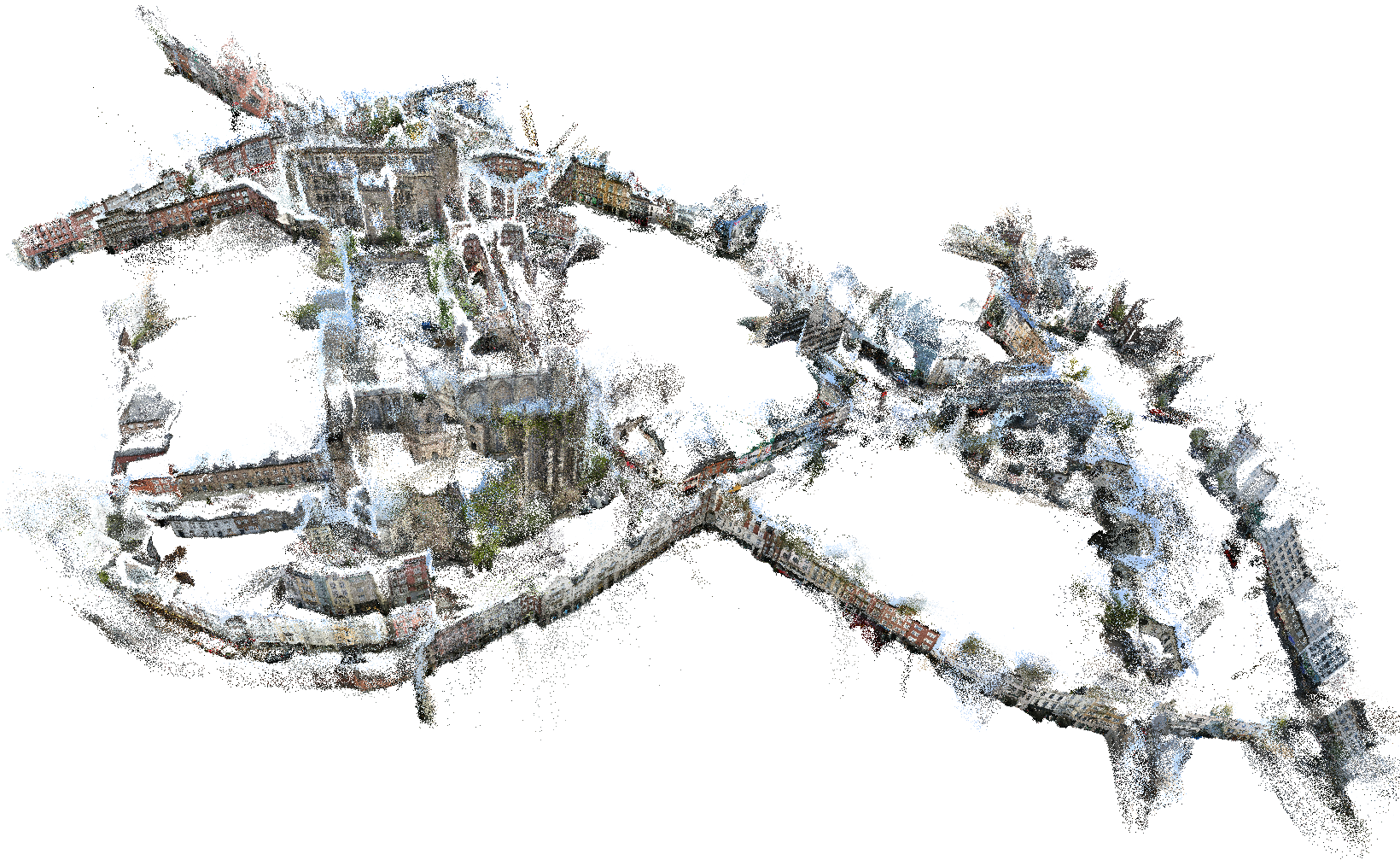} \includegraphics[width=0.37\linewidth]{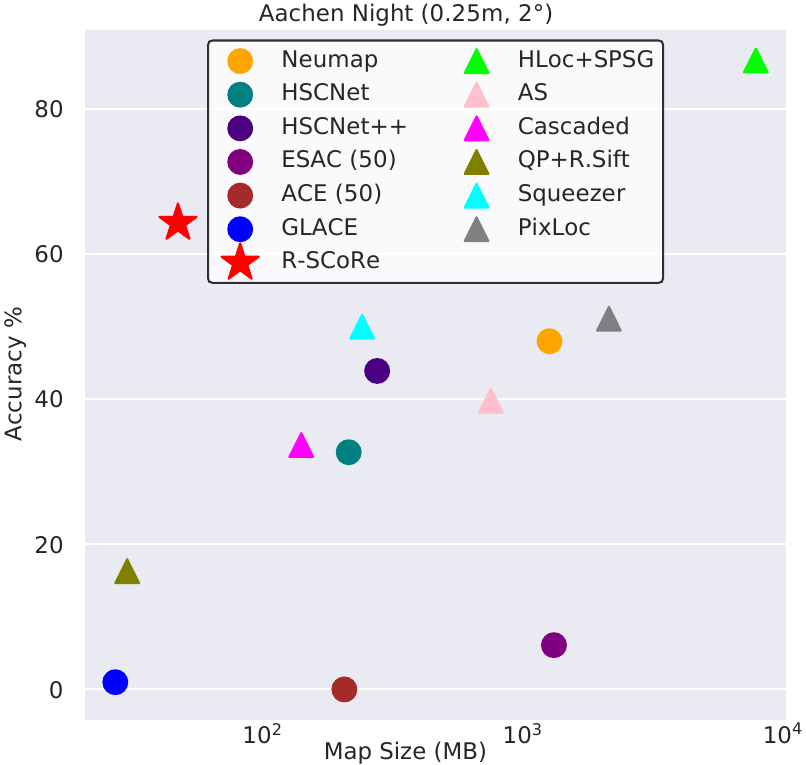}
    \captionof{figure}{\textbf{Robust Visual Localization with R-SCoRe.}
            \emph{Left}: Point cloud of Aachen reconstructed by R-SCoRe.
            \emph{Right}: On the large-scale Aachen Day-Night dataset~\cite{Sattler2012BMVC,Sattler2018CVPR} using only daytime training images, R-SCoRe achieves 64.3\% accuracy under the (0.25m, 2°) threshold for nighttime query images. It outperforms all previous SCR methods (circles) by a large margin. 
            With a small map size of only 47MB at a comparable accuracy, R-SCoRe is an attractive alternative to traditional methods (triangles). 
            }
    \label{fig:teaser_aachen_visualize}
\end{center}%
\vspace{0.2cm}

}]
\maketitle

{
    \renewcommand{\thefootnote}
        {\fnsymbol{footnote}}
        \footnotetext[1]{Corresponding author ({\tt fangjinhua.wang@inf.ethz.ch}).}
}

\begin{abstract}

    Learning-based visual localization methods that use scene coordinate regression~(SCR) offer the advantage of smaller map sizes.
    However, on datasets with complex illumination changes or image-level ambiguities, it remains a less robust alternative to feature matching methods.
    This work aims to close the gap. 
    We introduce a covisibility graph-based global encoding learning and data augmentation strategy, along with a depth-adjusted reprojection loss to facilitate implicit triangulation. 
    Additionally, we revisit the network architecture and local feature extraction module. 
    Our method achieves state-of-the-art on challenging large-scale datasets without relying on network ensembles or 3D supervision. 
    On Aachen Day-Night, we are 10$\times$ more accurate than previous SCR methods with similar map sizes 
    and require at least 5$\times$ smaller map sizes than any other SCR method while still delivering superior accuracy. Code is available at: \url{https://github.com/cvg/scrstudio}. %
    
\end{abstract}

\vspace*{-12pt}

\section{Introduction}

Visual localization is the task of estimating the 6-DoF pose of a camera in a known scene with a query image. 
It is a fundamental problem in computer vision, with applications in augmented reality, autonomous driving, and robotics.

Classical feature matching methods~\cite{sarlin2019coarse,sarlin2020superglue,detone2018superpoint,sattler2016efficient} have matured through years of research and now provide robust and accurate localization results. However, these methods typically require explicit 3D scene representations, where a large number of descriptors are stored, leading to substantial map sizes, especially for large-scale scenes. In contrast, pose regression~\cite{kendall2015posenet,Kendall2017GeometricLF,Brahmbhatt2018mapnet,Shavit2021MStransformer,turkoglu2021visual,WinkelbauerICRA21,zhou2020essnet,naseer2017deep} and scene coordinate regression (SCR)~\cite{brachmann2017dsac,brachmann2018dsacpp,brachmann2019esac,brachmann2021dsacstar,brachmann2023ace,dong2022visual,li2020hscnet,glace2024cvpr} aim to implicitly encode scene information in neural networks.

SCR methods follow a structure-based paradigm similar to feature matching,~\ie, estimating pose from 2D-3D correspondences but replacing explicit matching with regressing the correspondences directly. %
They are usually limited to small scenes~\cite{brachmann2023ace} and have yet to match feature matching methods in terms of accuracy and robustness. Recent advances~\cite{glace2024cvpr} extend SCR to large-scale scenes using a single model. However, its performance is still not on par with feature matching methods, especially in complex scenes with challenging illumination changes~\cite{Sattler2012BMVC,Sattler2018CVPR}.

In this work, we conduct a detailed analysis of the design principles behind the SCR framework, including local and global encoding, network architecture, and training strategies. Based on this analysis, we propose to revisit SCR to enhance the robustness and accuracy of SCR methods for large-scale visual localization tasks. 
As shown in \myfigref{fig:teaser_aachen_visualize}, our robust SCR (R-SCoRe) improves the night-time localization accuracy for SCR methods to 64.3\% under the~(0.25m, 2°) threshold on the Aachen Day-Night dataset, all with a map size of only 47MB. R-SCoRe significantly outperforms previous SCR methods and achieves accuracy comparable to feature matching techniques.

\mytabref{tab:methods} summarizes the practicability of R-SCoRe in complex large-scale scenes. While feature matching methods are also accurate, their map size can be prohibitively large, sometimes more than two orders of magnitude~\cite{sarlin2019coarse,sarlin2020superglue,detone2018superpoint}. 
Compared to SCR methods with similarly small map sizes~\cite{brachmann2023ace,glace2024cvpr}, R-SCoRe is at least one order of magnitude more accurate. 
While we still clearly outperform other SCR methods, we maintain fast inference and significantly smaller map sizes 
-- all without the need for scene-specific depth supervision.
Our contributions are as follows.
\begin{itemize}

    \item We propose learning a global encoding and performing data augmentation based on the covisibility graph. %
    To address the ambiguity of image retrieval features in complex large-scale scenes, we used multiple global hypotheses during testing.
    \item To unbias the network from neglecting near points, we introduce a depth-adjusted reprojection loss and show that this allows for accurate localization without scene-specific ground truth coordinate supervision. 
    \item To our knowledge, R-SCoRe is the first attempt of an SCR approach to achieve state-of-the-art performance on complex large-scale scenes without using an ensemble of networks or 3D model supervision.
\end{itemize}

\begin{table}[t]
	\centering
	
	\begin{tabular}{llcccc}
		\toprule
		& Methods &  Size & Time & Acc. & w/o Depth\\
		\midrule 
		FM&\cite{sarlin2019coarse,sarlin2020superglue,detone2018superpoint} & {\color{red}\xmark} & \cmark& \cmark & \cmark\\
              \midrule 
		PR&\cite{kendall2015posenet} & \cmark & \cmark & {\color{red}\xmark} & \cmark \\
  \midrule
		\multirow{4}{*}{SCR} &\cite{tang2023neumap}& {\color{red}\xmark} & {\color{red}\xmark}& \cmark & {\color{red}\xmark}\\
  &\cite{li2020hscnet,wang2024hscnet++,brachmann2019esac}& {\color{red}\xmark} & \cmark & \cmark & {\color{red}\xmark}\\
            &\cite{brachmann2018dsacpp,brachmann2023ace,glace2024cvpr} &              \cmark & \cmark & {\color{red}\xmark} & \cmark     \\
            &R-SCoRe& \cmark & \cmark & \cmark & \cmark  \\
		
		\bottomrule 
	\end{tabular}
\caption{\textbf{Comparison with other methods on complex large-scale scenes~\cite{Sattler2012BMVC,Sattler2018CVPR}.}  Feature matching (FM) methods are accurate but need a large map size. Pose regression (PR) methods are fast but less accurate. We maintain a small map size while achieving remarkable accuracy.
 }
\label{tab:methods}
\end{table}

\section{Related Work}

\customparagraph{Feature Matching}
Most state-of-the-art visual localization methods rely on feature matching \cite{sarlin2019coarse,sarlin2020superglue,detone2018superpoint,sattler2016efficient}. These methods typically adopt a structure-based paradigm, establishing 2D-3D correspondences between keypoints in a query image and 3D points in a scene. Camera pose is solved with geometric constraints, often a Perspective-n-Point (PnP) solver \cite{1217599,persson2018lambdap3p} within a RANSAC framework \cite{fischler1981random,barath2020magsac++,chum2003locally} to effectively manage outliers.
These methods commonly construct a Structure-from-Motion (SfM) map of the scene, which contains both 3D points and their descriptors \cite{schoenberger2016sfm,edstedt2024dedode,detone2018superpoint,RevaudNIPS19R2D2ReliableRepeatableDetectorsDescriptors,DusmanuCVPR19D2NetDeepLocalFeatures}. To efficiently establish 2D-3D matches, they often follow a two-level approach~\cite{sarlin2019coarse}. First, potentially relevant database images are retrieved using image retrieval techniques \cite{arandjelovic16netvlad,radenovic2018fine,zhu2023r2former}. This is followed by 2D-2D matching with the query image \cite{sarlin2019coarse,sarlin2020superglue,lindenberger2023lightglue}.

However, a significant limitation of these methods is the necessity to store all descriptor vectors of the 3D model, which can lead to storage challenges, particularly in large maps. To address this issue, various compression techniques have been proposed. These include reducing the number of 3D points \cite{6909460,7139575,li2010location,camposeco2019hybrid,yang2022scenesqueezer} or compressing descriptors \cite{dong2023learning,valenzuela2012dimensionality,ke2004pca,yang2022scenesqueezer,7410600,lynen2015get,jegou2010product,Laskar2024dpqed}.
Recently, several studies \cite{zhou2022gomatch,wang2024dgc,Panek2022meshloc} have proposed alternative approaches that eliminate the need for explicit descriptor storage. Instead, these methods advocate for direct matching against geometric representations, such as point clouds or meshes.

\customparagraph{Pose Regression}
These methods~\cite{kendall2015posenet,Kendall2017GeometricLF,Brahmbhatt2018mapnet,Shavit2021MStransformer,turkoglu2021visual,WinkelbauerICRA21,zhou2020essnet,naseer2017deep} directly estimates the camera pose of a query image with a neural network. %
However, they tend to struggle with generalization and often only achieve an accuracy similar to image retrieval methods~\cite{sattler2019limits}.

\begin{figure*}[th!]
    \centering
  \begin{subfigure}[b]{0.27\linewidth}
         \centering
         \includegraphics[width=\textwidth]{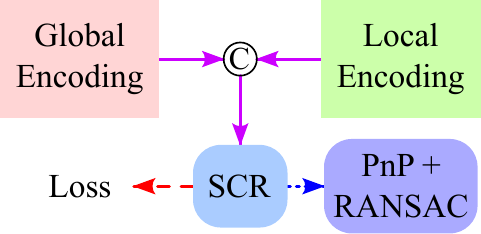}
         \caption{Workflow Overview}
         \label{fig:pipeline_overview}
     \end{subfigure}
\hfill
       \begin{subfigure}[b]{0.7\linewidth}
         \centering
         \includegraphics[width=\textwidth]{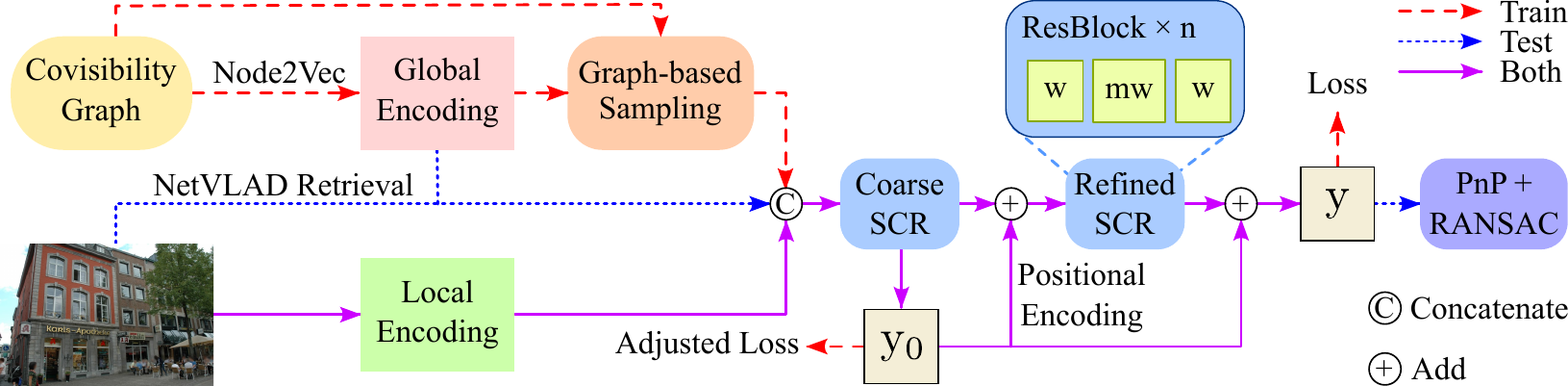}
         \caption{Detailed Pipeline}
         \label{fig:pipeline_detail}
     \end{subfigure}
    
    \mycaption{R-SCoRe pipeline.}{
(a) Following the SCR workflow in~\cite{glace2024cvpr}, we concatenate patch-level local encodings with image-level global encodings as input to a scene-specific MLP.  
(b) We learn contrastive global encodings from the covisibility graph using Node2Vec~\cite{node2vec-kdd2016}.
During training, global encodings are sampled from neighboring nodes for data augmentation. During inference, we retrieve global encodings from the $k$ nearest training images via NetVLAD~\cite{arandjelovic16netvlad} as hypotheses and select the one yielding the most RANSAC inliers. We enhance the SCR MLP with a refinement module and introduce a depth-adjusted reprojection loss to reduce bias toward distant points. 
    }
    \label{fig:our_pipeline}
\end{figure*}

\customparagraph{Scene Coordinate Regression} 
Following a similar structure-based localization paradigm as feature matching methods, SCR methods~\cite{brachmann2016,cavallari2017fly,Cavallari2019cascade,shotton2013scene,valentin2015cvpr,brachmann2017dsac,brachmann2018dsacpp,brachmann2019esac,brachmann2021dsacstar,brachmann2023ace,dong2022visual,li2020hscnet,wang2024hscnet++,focustune,glace2024cvpr} regress 2D-3D correspondences between the query image and the scene, and estimate the camera pose using geometric constraints. Though SCR methods are more accurate than pose regression methods, they usually still struggle with large-scale scenes~\cite{brachmann2023ace}.

Recently, several approaches have been proposed to improve the scalability and performance of SCR in large-scale scenes. These methods often rely on ground truth 3D coordinates and aim to handle large scenes by dividing them into smaller segments, such as spatial regions~\cite{brachmann2019esac}, voxels~\cite{tang2023neumap}, or hierarchical clusters~\cite{li2020hscnet,wang2024hscnet++}. 
Recent advancements have explored alternatives that do not require ground truth 3D supervision. For example, ACE~\cite{brachmann2023ace} uses reprojection loss only for training. GLACE~\cite{glace2024cvpr} further introduces a global encoding mechanism that eliminates the need for scene segmentation.
Despite these improvements, these methods still encounter limitations under challenging conditions, such as significant changes in illumination.

\section{Method}

Our Scene Coordinate Regression (SCR) workflow is depicted in \myfigref{fig:pipeline_overview}.
During training, we have access to a covisibility graph to learn global encodings which are concatenated to the local encodings. During inference, we retrieve global encoding hypotheses from the $k$ nearest training images and predict 2D-3D correspondences: we run PnP for each hypothesis and select the one yielding the most RANSAC inliers. 
\myfigref{fig:pipeline_detail} illustrates our detailed R-SCoRe pipeline, where we split SCR into coarse and refinement blocks and introduce covisibility-based global encoding and data augmentation techniques.

\subsection{Preliminaries}

\customparagraph{Visual Localization}
Given a test image $I_{\text{test}}$ with known intrinsic $\mat{K}_{\text{test}}$, the goal is to estimate its extrinsic,
i.e., the rigid transformation $[\mat{R}_{\text{test}}|\vec{t}_{\text{test}}]$ from world coordinate to camera coordinate. 
The scene is typically given by a set of training images $I_{\text{train}}$ with known ground truth poses $[\mat{R}_{\text{train}}|\vec{t}_{\text{train}}]$ and intrinsics $\mat{K}_{\text{train}}$.

\customparagraph{Scene Coordinate Regression}
The SCR pipeline employs a neural network $f$ to directly regress the 3D coordinate $y = f(\mathbf{F}(x))$ for each 2D keypoint $x$ with feature $\mathbf{F}(x)$. 
Without the need to store large point clouds with descriptors, SCR methods implicitly represent the scene with a neural network, which usually results in a smaller map size.

\customparagraph{Scalable SCR without 3D ground truth}
Recent advances~\cite{brachmann2023ace,glace2024cvpr} allow SCR to scale to large scenes without scene-specific 3D supervision. 
To reduce ambiguities in large scenes, GLACE~\cite{glace2024cvpr} (\myfigref{fig:pipeline_overview}) concatenates a local patch-level encoding with an image-level global encoding as the keypoint feature \( \mathbf{F}(x) \). The local encoder is a pretrained DSAC*~\cite{brachmann2021dsacstar} backbone, following~\cite{brachmann2023ace}. 
The global encoder uses a pretrained image retrieval model~\cite{zhu2023r2former}
with Gaussian noise augmentation to prevent overfitting to trivial solutions, \cf~\cite{glace2024cvpr}. 
To accelerate training, all features are precomputed and buffered in GPU memory, from which a random sample is drawn for each batch.

Without ground truth scene coordinates, the output 3D point $y$ is reprojected with the ground truth pose $\mat{R}, t$ and intrinsics $\mat{K}$, and compared to the keypoint location $x$ in 2D:
\begin{equation}\label{eq:reprojection}
    e_2(x, y) = ||x - \pi(\mat{K} (\mat{R} y + \vec{t}))||_2,
\end{equation}
where $\pi$ converts homogeneous to Cartesian coordinates.

Instead of explicitly grouping corresponding observations into tracks, this underconstrained supervision is applied to each independent prediction. Prior works~\cite{glace2024cvpr,focustune} suggest that implicit triangulation can still occur as the network tends to produce similar outputs for similar inputs.

The reprojection error is fed into a dynamic robust loss, \cf ACE~\cite{brachmann2023ace}, to focus on points accurately regressed:
\begin{equation}\label{eq:robust}
l_{\text{dynamic}}(e_2(x, y)) = \tau(t) \rho \left(\frac{e_2(x, y)}{\tau(t)}\right),
\end{equation} 
where ACE~\cite{brachmann2023ace} uses $\tanh$ as robust loss ($\rho:=\tanh$).
Based on the relative training time $t\in[0,1]$, the bandwidth $\tau(t)$ is adjusted dynamically during training: 
\begin{equation}\label{eq:schedule} 
\tau(t) = \sqrt{1 - t^2}  \tau_{\max} + \tau_{\min}. 
\end{equation}
Keypoints $x$ whose regressed 3D point $y$ fall outside the valid frustum are penalized differently. 
Valid points are defined to lie within a valid depth range $[d_{\min}, d_{\max}]$ in front of the camera. 
Further, their reprojection error $e_2(x, y)$ must be smaller than a threshold $e_{\max}$. 
For invalid points, we penalize their distance to a pseudo ground truth point $\bar{y}$: 
\begin{equation} 
l_{\text{invalid}}(y) = ||y - \bar{y}||_2, 
\end{equation} 
where $\bar{y}$ is computed by the inverse projection of the pixel $x$ using
a fixed target depth $d_{\text{target}}$.

\subsection{Network Architecture}
We adopt the MLP architecture and position decoder from GLACE~\cite{glace2024cvpr}, scaling the network width with scene size. As illustrated in \myfigref{fig:pipeline_detail}, we also introduce a refinement module at the end of the network, which adjusts the final output \(y\) by predicting an offset from the intermediate prediction \(y_0\). The coarse coordinate \(y_0\) is reintroduced into the refinement module through positional encoding~\cite{tancik2020fourfeat,mildenhall2020nerf} using sine and cosine functions with periods ranging from 0.5 to 2048, which is added to the intermediate feature. 
This empirically improves training stability and allows the network to achieve lower training reprojection errors more rapidly.

\subsection{Input Encoding}
\customparagraph{Analysis}  
In implicit triangulation, reprojection constraints are grouped based on input similarity. Therefore, the desired properties of input encodings are as follows: positive pairs observing the same points should produce similar features, while negative pairs observing distinct points should yield clearly distinguishable features.  Additionally, it is preferable for the encodings to be low-dimensional to minimize memory requirements. %

\customparagraph{Local Encoding}
At the local patch level, features should differentiate observations of the same point from those of different points.
The requirement aligns with the properties of local descriptors used in traditional feature matching, suggesting that we can directly leverage their local feature extractors. 
We investigate pretrained feature extractors for both dense and sparse matching methods, such as LoFTR~\cite{sun2021loftr} and Dedode~\cite{edstedt2024dedode}. 
To lower memory consumption during training, we apply PCA to all the features from the training dataset, reducing their dimensionality while retaining most of the variance. 
We experimentally observe that reducing the dimensionality to 128 dimensions preserves over 90\% of the variance on various datasets~\cite{Sattler2012BMVC,Sattler2018CVPR,lee2021indoorlocdataset}. %

\begin{figure}[t]
    \centering
    \begin{subfigure}[b]{\linewidth}
   \centering
   \includegraphics[width=\linewidth]{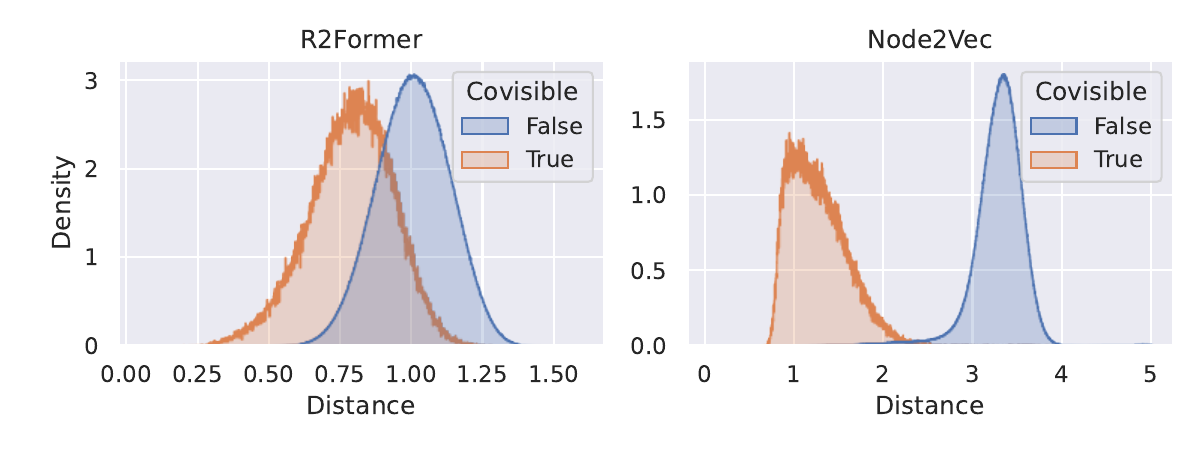}
   \caption{Distribution of feature distance for covisible and non-covisible pairs.}
    \end{subfigure}
    \vfill
    \begin{subfigure}[b]{\linewidth}
   \centering
   \includegraphics[width=0.75\linewidth]{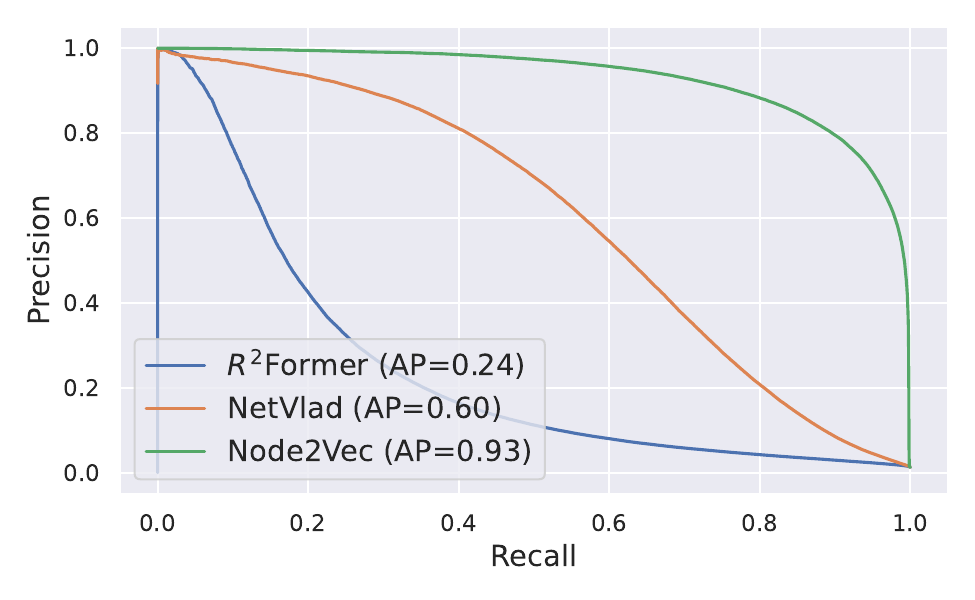}
   \caption{Precision Recall Curve of predicting covisibility by feature distance.}
    \end{subfigure}
    \mycaption{Comparison of global encodings.}{
    Aligning the learning of global encodings with the covisibility graph topology (Node2Vec~\cite{node2vec-kdd2016}) helps distinguish covisible and non-covisible pairs (a) and predict covisibility by feature distance (b).
    }
   \label{fig:global_encoding_stats}
\end{figure}

\customparagraph{Covisibility Graph Based Global Encoding}
Image-level global features should distinguish between covisible and non-covisible image pairs, i.e. whether the images are viewing the same part of the scene, to resolve ambiguities in local encodings. 
Although global encodings with image-level receptive fields can help, they may still be insufficient to resolve ambiguities in complex environments, as shown in \myfigref{fig:global_encoding_stats}. 
This limitation can lead to imperfect grouping of reprojection constraints during training, thereby impairing the effectiveness of implicit triangulation. 
Furthermore, we point out that the learned SCR function may lack (Lipschitz) smoothness~\cite{khromovS24} \wrt the global encodings if adapted naively, \eg minor variations in the global encoding can result in significant shifts in corresponding 3D points and consequently reduce generalization at test time.

To address these issues, we propose to directly learn embeddings aligned with the covisibility graph's topology using Node2Vec~\cite{node2vec-kdd2016}, which samples sequences with weighted random walks and optimizes node embeddings with a Skip-gram~\cite{mikolov2013efficient} objective. For training images, the covisibility graph is easily available. It can be estimated from the frustum overlap of ground truth poses (see the supplementary for more details). At test time, however, covisibility information is unknown, so we propose generating multiple global encoding hypotheses by retrieving the nearest training images using NetVLAD~\cite{arandjelovic16netvlad} features. The global encoding of each retrieved image serves as a hypothesis, and we select the hypothesis yielding the maximum RANSAC inliers for the final localization result.

This approach effectively decouples training-time and test-time ambiguities: during training, the network focuses on learning scene structure without ambiguity, while at test time, multiple hypotheses enable the resolution of complex, often multimodal ambiguities.

\customparagraph{Covisibility Graph Based Data Augmentation}
 Our Covisibility Graph Encoding effectively learns a low-ambiguity global encoding. However, data augmentation is still necessary to prevent the network from distinguishing covisible pairs based on distinct global encodings. Instead of simply adding isotropic Gaussian noise~\cite{glace2024cvpr}, we introduce a graph-based data augmentation strategy. In this approach, rather than applying isotropic noise, we randomly replace an image's global encoding with that of a neighboring image from the covisibility graph. Specifically, with probability $p=0.5$, the current image's global encoding is retained, while with probability $1 - p$, it is replaced by the global encoding of a randomly sampled neighboring image.

\subsection{Output Supervision}

\begin{figure}[t!]
    \centering
    \includegraphics[width=\linewidth]{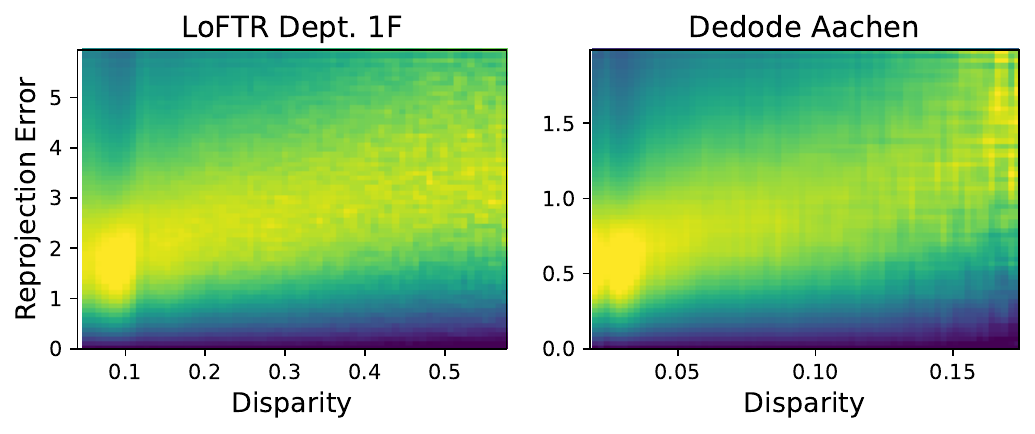}
    \mycaption{Statistics of reprojection error for points with different depths.}{ 
The kernel density estimation (KDE) of reprojection error distribution conditioned on disparity from SCR models trained with various local encodings across different datasets. 
We observe that far points (low disparity) exhibit a lower reprojection error. (Detector-free LoFTR~\cite{sun2021loftr} with an 8 $\times$ downsampled output has a larger 2D keypoint error than Detector-based Dedode~\cite{edstedt2024dedode}.)
}

    \label{fig:reperr_disparity}
\end{figure}

\customparagraph{Depth Bias in Reprojection Loss} 
\myfigref{fig:reperr_disparity} displays statistics collected from training SCR models with various local encodings across different datasets. 
It indicates that points closer to the camera empirically exhibit higher reprojection errors compared to distant points, hence we observe a bias toward distant points.
This bias is magnified by training with the robust loss~\myeqref{eq:robust}, 
as the supervision signal tends to neglect (nearby) points with higher reprojection errors (\cf~\myfigref{fig:supervision_median_depth}). Assuming the training time distribution of camera poses to be representative for testing, 
regressing near points fewer and less accurately during testing diminishes the positional localization accuracy.
We conjecture that to facilitate implicit triangulation for near points, a higher reprojection error should be allowed to compensate for 
the reprojection of nearby points being more sensitive to pose variations and coordinate inaccuracies.

\customparagraph{Depth Adjusted Reprojection Error} 
We propose normalizing the reprojection error in \myeqref{eq:reprojection} based on the depth of the predicted scene coordinate. %
Specifically, the observation standard deviation, $\sigma_o$, is defined as:
\begin{equation}
    \sigma_o = \sqrt{\left(\frac{\sigma_3}{d}\right)^2 + \sigma_2^2},
\end{equation}
where we denote the variance of the noise of the 2D observations by $\sigma_2^2$. The variance of the 3D prediction is denoted by $\sigma_3^2$ and by \(d\) the depth of the point. 
The reprojection error is then adjusted accordingly:
\begin{equation}\label{eq:adjusted_reprojection}
e_3(x, y) = \frac{e_2(x,y)}{\sigma_o} = \frac{e_2(x,y)}{\sigma_2}\sqrt{\frac{d^2}{d^2 + \left(\frac{\sigma_3}{\sigma_2}\right)^2}}.
\end{equation}

\customparagraph{Selective Application of Depth Adjustment}  
The bias towards distant points may sometimes actually be beneficial, as underconstrained points can be pushed farther along the ray, making them easier to identify as outliers during test time. 
Therefore, we apply our depth-adjusted reprojection loss only to the intermediate coarse scene coordinate output $y_0$ during training~(\myfigref{fig:pipeline_detail}), and retain the original reprojection loss for the final output. 
To mitigate the concentration of the supervision signal on points with low projection error, 
we replace $\tanh$ with the Geman-McClure~\cite{barron2019generalrobust} robust loss function which has a heavier tail than $\tanh$: %
\begin{equation}
    \rho(x)=\frac{9x^2}{9x^2+4}.
\end{equation}

In order to guide the convergence of the regressed points, $y$, by the intermediate output $y_0$ (affected by the normalized loss from \myeqref{eq:adjusted_reprojection}), 
we also apply a consistency loss at the beginning of the training: %
\begin{equation}
    l_{\text{consistency}} = \lambda(t) ||y - y_0||_2, 
    \label{eq:consistency_loss}
\end{equation}
where $\lambda(t)$ is a dynamic weight that decreases to 0 in a cosine schedule during the first 50\% of training time.
\begin{equation}
    \lambda(t) = \begin{cases}
       \frac{1}{2} \left(1 + \cos 2\pi t\right), & \text{if } t \in [0, 0.5] \\
        0, & \text{otherwise}
    \end{cases},
\end{equation}
where $t$ is the relative training time.

\customparagraph{Optional Depth Supervision}
When depth is available, we can also benefit from direct depth supervision. 
The depth does not need to be accurate since we mainly use the depth for initialization. 
Specifically, we simply replace the consistency loss between intermediate and final output in \myeqref{eq:consistency_loss} with a ground truth coordinate supervision loss:
\begin{equation}
    l_{\text{depth}} = \lambda(t) \left( ||y - \bar{y}||_2 + ||y_0 - \bar{y}||_2 \right), 
\end{equation}
where $\bar{y}$ is the pseudo ground truth computed by the inverse projection of the pixel given the depth, pose, and intrinsic.

\begin{table*}[h!]
    \centering
    \begin{tabular}{l|c|c|*{3}{c}|*{3}{c}}
        \hline
        Methods                                                                    & w/o Depth & Size   & \multicolumn{3}{c|}{Aachen Day} & \multicolumn{3}{c}{Aachen Night}                             \\
        \hline
        HLoc+SPSG~\cite{sarlin2019coarse,sarlin2020superglue,detone2018superpoint} & Yes       & 7.82GB & 89.6                            & 95.4                             & 98.8 & 86.7 & 93.9 & 100  \\
        AS~\cite{sattler2016efficient}                                             & Yes       & 750MB  & 85.3                            & 92.2                             & 97.9 & 39.8 & 49.0 & 64.3 \\
        Cascaded~\cite{cheng2019cascaded}                                          & Yes       & 140MB  & 76.7                            & 88.6                             & 95.8 & 33.7 & 48.0 & 62.2 \\
        QP+R.Sift~\cite{mera2020efficient}                                         & Yes       & 30MB   & 62.6                            & 76.3                             & 84.7 & 16.3 & 18.4 & 24.5 \\
        Squeezer~\cite{yang2022scenesqueezer}                                      & Yes       & 240MB  & 75.5                            & 89.7                             & 96.2 & 50.0 & 67.3 & 78.6 \\
        \hline
        PixLoc~\cite{sarlin2021back}                                               & Yes       & 2.13GB & 64.3                            & 69.3                             & 77.4 & 51.1 & 55.1 & 67.3 \\
        \hline
        Neumap~\cite{tang2023neumap}                                               & No        & 1.26GB & 80.8                            & 90.9                             & 95.6 & 48.0 & 67.3 & 87.8 \\
        HSCNet~\cite{li2020hscnet}                                                 & No        & 213MB  & 71.1                            & 81.9                             & 91.7 & 32.7 & 43.9 & 65.3 \\
        HSCNet++~\cite{wang2024hscnet++}                                           & No        & 274MB  & 72.7                            & 81.6                             & 91.4 & 43.9 & 57.1 & 76.5 \\
        ESAC ($\times50$)~\cite{brachmann2019esac}                                 & No        & 1.31GB & 42.6                            & 59.6                             & 75.5 & 6.1  & 10.2 & 18.4 \\
        ACE ($\times50$)~\cite{brachmann2023ace}                                   & Yes       & 205MB  & 6.9                             & 17.2                             & 50.0 & 0.0  & 1.0  & 5.1  \\
        GLACE~\cite{glace2024cvpr}                                                 & Yes       & 27MB   & 8.6                             & 20.8                             & 64.0 & 1.0  & 1.0  & 17.3 \\
        \hline

        R-SCoRe (Dedode~\cite{edstedt2024dedode})                                     & Yes       & 47MB &  74.8  &  86.9  &  96.4 & 64.3  &  89.8  &  96.9 \\
        ~~+ Depth                                                                  & No        & 47MB   & 79.0  &  88.5  &  96.4 & 66.3  &  89.8  &  96.9 \\
        \hline
    \end{tabular}
    \mycaption{Aachen Day-Night evaluation.}{The map size and percentages of query images within three thresholds: (0.25m, 2°), (0.5m, 5°), and (5m, 10°) and are reported. We report our results with Dedode~\cite{edstedt2024dedode} local encoding and optional depth supervision. Feature matching (FM) methods~\cite{sarlin2019coarse,sarlin2020superglue,detone2018superpoint} are more accurate, but the map size is large. R-SCoRe achieves comparable accuracy with a small map size.
    }
    \label{tab:aachen}
\end{table*}

\section{Experiments}

\subsection{Datasets}
We use the Aachen Day-Night~\cite{Sattler2012BMVC,Sattler2018CVPR} and the Hyundai Department Store dataset~\cite{lee2021indoorlocdataset} to evaluate R-SCoRe on complex large-scale indoor and outdoor scenes.

\customparagraph{Aachen Day-Night}
It is a large-scale benchmark for outdoor visual localization, covering the historic inner city of Aachen, Germany, over an area of approximately 6 $km^2$. It presents significant challenges due to varying illumination conditions, especially between day and night. The dataset includes 4,328 daytime images for training, along with 824 daytime query images and 98
nighttime query images.

\customparagraph{Hyundai Department Store}
It is a large-scale indoor visual localization benchmark, covering three floors of a department store. Each floor consists of multiple sequences captured over four months, spanning an area of approximately 10,000 $m^2$. It presents challenges beyond its large scale, including dynamic objects, illumination changes, and textureless regions. B1 is particularly challenging as the training images are captured under low-lighting conditions, while the query images are brightly illuminated. The dataset includes 44,283 training images and 5,927 test images.

\subsection{Benchmark Results}

\customparagraph{Aachen Day-Night}
As shown in \mytabref{tab:aachen}, R-SCoRe enhances the performance of scene coordinate regression (SCR) based methods~\cite{brachmann2019esac,glace2024cvpr,wang2024hscnet++,li2020hscnet}, achieving competitive results with a single low-map-size model without the need for scene-specific depth supervision.
While R-SCoRe is competitive with the best performing method, HLoc~\cite{sarlin2019coarse,sarlin2020superglue,detone2018superpoint}, we forfeit some ground at highest accuracy. However, R-SCoRe demands 170$\times$ less memory to store the map.
This huge gap could already render SCR based methods as an attractive alternative for some applications.
Most other feature based methods (FM) also deliver significantly larger maps. While delivering comparable performance for the Aachen Day dataset, they all fall behind R-SCoRe on the Aachen Night dataset.
The only FM method~\cite{mera2020efficient} with a comparable map size is outperformed on all metrics, \eg \cite{mera2020efficient} is 4-5$\times$ worse in accuracy on the night dataset.
Compared to other SCR methods that work without depth supervision \cite{brachmann2023ace,glace2024cvpr} R-SCoRe is 10$\times$ superior in accuracy. %
The so far most accuracte SCR based method \cite{tang2023neumap} produces large maps (27$\times$ larger) and is prohibitively slow in inference.
The next most accurate SCR method, \cite{wang2024hscnet++} is outperformed by 46\% at night and highest threshold, while R-SCoRe maintains a 6$\times$ smaller map size -- without the need for depth supervision.
We observe a small gain in performance if we utilize depth for supervision of R-SCoRe.

\begin{table*}[h!]
    \centering
    \begin{tabular}{l|c|c|c}
        \hline
        Methods                                                                                 & Dept. 1F Test                             & Dept. 4F Test                        & Dept. B1  Test                 \\
        \hline
        HLoc+D2-Net~\cite{sarlin2019coarse,DusmanuCVPR19D2NetDeepLocalFeatures}                 & (78.0 / 82.8 / 88.0) /        398GB       & (84.2 / 89.8 / 92.0)   / 183GB       & (73.7 / 79.3 / 87.2) /  505GB  \\
        HLoc+R2D2~\cite{sarlin2019coarse,RevaudNIPS19R2D2ReliableRepeatableDetectorsDescriptors} & (80.6 / 84.3 / 89.4) /   166GB            & (85.3 / 91.0 / 93.1)   / 76GB       & (75.2 / 80.3 / 87.6) /  210GB  \\
        \hline
        PoseNet~\cite{kendall2015posenet}                                                       & (0.0 / 0.0 / 0.4) /                  41MB & (0.0 / 0.0 / 0.1)   /         41MB   & (0.0 / 0.0 / 0.1) /     41MB   \\
        \hline
        ESAC ($\times50$)~\cite{brachmann2019esac}                                              & (43.3 / 66.3 / 77.0) /1.4GB               & (45.2 / 62.5 / 73.1) / 1.4GB         & (3.5 / 8.2 / 12.6) /     1.4GB \\
        ACE ($\times50$)~\cite{brachmann2023ace}                                                & (14.1 / 54.4 / 75.5) / 205MB              & (27.3 / 70.9 / 84.1) / 205MB         & (2.7 / 14.4 / 29.3) / 205MB    \\
        GLACE~\cite{glace2024cvpr}                                                              & (5.6 / 21.3 / 48.6) / 42MB                & (8.4 / 29.8 / 51.6) / 42MB           & (0.9 / 4.4 / 11.9) /  42MB     \\
        \hline
        R-SCoRe (LoFTR$^*$~\cite{sun2021loftr})                                                    & (63.2 / 82.4 / 92.4)  /  127MB            & (62.2 / 82.7 / 90.9) / 50MB          & (26.9 / 50.7 / 69.6) /   130MB \\
        ~~+ Depth                                                                               & (67.3 / 84.5 / 92.6) /  127MB             & (70.5 / 87.0 / 92.9) / 50MB          & (30.8 / 53.7 / 72.7) /  130MB              \\
        R-SCoRe (Dedode~\cite{edstedt2024dedode})                                                  & (61.4 / 80.2 / 90.9) /  127MB             & (60.2 / 79.3 / 87.9) / 50MB          & (60.1 / 77.3 / 89.6)  /  130MB \\
        ~~+ Depth                                                                               & (63.9 / 83.3 / 90.8) /  127MB             & (76.7 / 89.3 / 93.0) / 50MB          & (61.5 / 77.6 / 88.8)  /  130MB \\
        \hline
    \end{tabular}

    \mycaption{Hyundai Department Store Test Set evaluation.}{The percentages of query images within three thresholds: (0.1m, 1°), (0.25m, 2°), and (1m, 5°) and the map size are reported.
    R-SCoRe achieves competitive accuracy with a small map size.
    $^*$We use LoFTR~\cite{sun2021loftr} outdoor, trained on MegaDepth~\cite{li2018megadepth}, instead of the indoor model trained on ScanNet~\cite{dai2017scannet} for the B1 scene with strong illumination change.
    }
    \label{tab:test_hyundai}
\end{table*}

\customparagraph{Hyundai Department Store}
R-SCoRe again significantly outperforms the SCR based methods~\cite{brachmann2019esac,glace2024cvpr}, including recent ensemble networks~\cite{brachmann2019esac,brachmann2023ace}, see \mytabref{tab:test_hyundai}.
Compared to the state-of-the-art feature matching based localization~\cite{sarlin2019coarse} we achieve competitive results with a single low-map-size model and 
forfeit some ground at the highest accuracy threshold.
However, our model is at least three orders of magnitude smaller for either feature~\cite{DusmanuCVPR19D2NetDeepLocalFeatures,RevaudNIPS19R2D2ReliableRepeatableDetectorsDescriptors} incorporated into~\cite{sarlin2019coarse}, which can be a valuable advantage in practice.
Recall that depth supervision is not necessary for R-SCoRe, but if available, it can also enhance performance further.
The B1 scene exhibits strong illumination changes and we observe significantly better performance when using local encodings from Dedode instead of LoFTR~\cite{sun2021loftr}.

\subsection{Implementation Details}
Most hyperparameters follow default values~\cite{glace2024cvpr} and extensive tuning is not performed, as we empirically find the approach remains robust within a reasonable range, apart from the trade-off between network size and performance.

\customparagraph{Input Encodings}
For local encodings, we perform PCA to reduce their dimensionality to 128.
The global encodings are represented in 256 dimensions, consistent with the \(R^2\)Former~\cite{zhu2023r2former} feature dimension used in GLACE~\cite{glace2024cvpr} for fair comparison.  We estimate the covisibility graph based on camera poses, using a weighted frustum overlap method (details provided in the supplementary materials), with a maximum viewing frustum depth of \(d_{\text{v}} = 50\) for outdoor scenes and \(d_{\text{v}} = 8\) for indoor scenes.

\customparagraph{Output Supervision}
The supervision uses a dynamic robust loss bandwidth strategy inspired by ACE~\cite{brachmann2023ace}. For coarse intermediate outputs, the parameters, see Eq. \eqref{eq:schedule}, are set to \( \tau_{\text{min}} = 1 \) and \( \tau_{\text{max}} = 50 \). In contrast, \( \tau_{\text{max}} = 25 \) is used for the final output, which allows the refinement layer to focus on the most accurate predictions while the initial layers do not ignore the optimization of relatively inaccurate predictions.
Fixing \( \sigma_2 = 1 \) in the depth-adjusted reprojection loss, Eq.~\eqref{eq:adjusted_reprojection}, allows us to control the behavior by adjusting \( \tau \) and \( \frac{\sigma_3}{\sigma_2} \).
For indoor scenes, \( \sigma_3 = 3 \) is applied, while \( \sigma_3 = 8 \) is used for outdoor scenes to account for different scales.
We perform optional depth supervision using depth images rendered from the 3D model for the Hyundai Department Store dataset and Multi-View Stereo depth maps for the Aachen Day-Night dataset.

\customparagraph{Network Architecture}
We adopt the MLP architecture and position decoder from GLACE~\cite{glace2024cvpr} with expansion ratio $m = 2$ for the MLP, and 50 clusters for the position decoder. %
With MLP width   
$
w\!=\!256\left\lceil\sqrt{n/1000}\right\rceil
$ for $n$ training images, we scale the parameter count proportionally.

\customparagraph{Training}
We found that adopting the optimization settings from ACE Zero~\cite{brachmann2024acezero} enhances both stability and convergence speed compared to the original ACE~\cite{brachmann2023ace}. Specifically, we reduce the warmup ratio of the one-cycle learning rate schedule~\cite{smith2019cycliclr} from 0.25 to 0.04 and lower the peak learning rate from \(5 \times 10^{-3}\) to \(3 \times 10^{-3}\).
For our evaluation we adopt similar training parameters to GLACE~\cite{glace2024cvpr}, including a local feature buffer size of 128M, a large batch size of 320K and a training duration of 100k iterations.

\customparagraph{Testing}
At test time, we retrieve the 10 nearest training images with NetVLAD~\cite{arandjelovic16netvlad}. The global encoding and retrieval features for training images are precomputed and compressed using Product Quantization~\cite{jegou2010product}. For final pose estimation, we utilize PoseLib~\cite{poselib} with a maximum reprojection error of 10 pixels and up to 10,000 RANSAC iterations. 
On our PC (NVIDIA RTX 2080 GPU \& Intel i7-9700K CPU), the average inference time for a $640\!\times\!480$ query image is 140 to 270 ms in total.%
\begin{itemize}
\item Global: NetVLAD (20ms), Retrieval ($<$1ms)  
\item Local: LoFTR (7ms) or DeDoDe (50ms) 
\item MLP: w = 768 (70ms) or 1280 (160ms)
\item Pose Solving: 40ms  
\end{itemize}

\subsection{Ablation Study Results}

\begin{table}[t]
    \centering
    \begin{tabular}{l|ccc|ccc}
        \hline
                                        & \multicolumn{3}{c|}{Dept. 1F Val} & \multicolumn{3}{c}{Dept. B1 Val}                             \\
        \hline

        ACE~\cite{brachmann2023ace}     & 68.7                              & 87.5                             & 95.9 &  14.1 & 28.3 & 45.8 \\
        LoFTR$^*$~\cite{sun2021loftr}       & 72.3                              & 88.7                             & 95.5 &  29.4 & 51.3 & 69.6 \\
        Dedode~\cite{edstedt2024dedode} & 70.6                              & 86.6                             & 95.5 &  57.7 & 74.7 & 86.7 \\
        \hline
    \end{tabular}
    \mycaption{Ablation study of local encoders.}{Accuracy at (0.1m, 1°), (0.25m, 2°), and (1m, 5°) thresholds are reported. Utilizing pretrained, off-the-shelf feature extractors improves the performance, especially under challenging conditions (B1).}
    \label{tab:local_hyundai}
\end{table}

\begin{table}[t]
    \centering
    \begin{tabular}{l|rrr}
        \hline
                                                         & \multicolumn{3}{c}{Dept. 1F Val}               \\
        \hline
        ~~$R^2$Former~\cite{zhu2023r2former} w/ Gaussian & 34.1                             & 60.1 & 78.3 \\
        + Multi Hypotheses                               & 42.1                             & 74.5 & 92.2 \\
        + Covis Augmentation                             & 62.0                             & 83.8 & 94.8 \\
        + Covis Encoding                                 & 72.3                             & 88.7 & 95.5 \\
        \hline
    \end{tabular}
    \mycaption{Ablation study of global encodings.}{
        We experiment with using multiple hypotheses at test time, applying covisibility graph-based data augmentation during training, and learning global encodings directly from the covisibility graph.
        Accuracy at (0.1m, 1°), (0.25m, 2°), and (1m, 5°) thresholds.
        }
    \label{tab:ablation_global}
\end{table}

In our ablation studies, we investigate the impact of the different components in R-SCoRe. %
We evaluate on the validation split for the Hyundai Department Store dataset. %
Since the Aachen Day-Night dataset does not provide a validation split, we evaluate on the test set.

\customparagraph{Local Encoding}
As shown in \mytabref{tab:local_hyundai}, for large scale indoor scenes with small illumination changes, alternative off-the-shelf local feature extractors~\cite{edstedt2024dedode,sun2021loftr} achieve similar or even superior performance compared to the original ACE~\cite{brachmann2023ace}. Note that this finding contradicts earlier investigations~\cite{brachmann2023ace} that prefer a specifically trained backbone in their work. Additionally, local descriptors trained on MegaDepth~\cite{li2018megadepth}, especially Dedode~\cite{edstedt2024dedode}, demonstrate greater robustness in scenes with significant illumination changes.

\customparagraph{Global Encoding}
Retrieving global encodings from training images avoids the domain gap. %
Better retrieval method and multiple hypotheses verification can help resolve ambiguities.
Without retraining (\mytabref{tab:ablation_global}), utilizing multiple global hypotheses at test time (\emph{+ Multi Hypotheses}) results in a direct performance improvement in complex scenes.
The performance improves significantly, once we incorporate our covisibility graph-based data augmentation during training.
In particular, we replace isotropic Gaussian noise~\cite{glace2024cvpr} with our more precise covisibility-based technique (\emph{+ Covis Augmentation}).
Finally, learning the global encoding directly from the covisibility graph (\emph{+ Covis Encoding}) reduces the interference between non-covisible training images and thereby facilitates implicit triangulation, especially in indoor scenes with significant ambiguity.

Finally, we also explore the effect of computing the covisibility graph via feature matching~\cite{sarlin2020superglue,detone2018superpoint}. As shown in \mytabref{tab:covgraph_quality}, using a more accurate graph yields no significant improvement, indicating that R-SCoRe is robust to the quality of the covisibility graph. Therefore, our simple frustum overlap-based graph is sufficient for effective performance.

\begin{table}[t]
    \centering
    \begin{tabular}{l|*{3}{r}|*{3}{r}}
        \hline
                   & \multicolumn{3}{c|}{Aachen Day} & \multicolumn{3}{c}{Aachen Night}                             \\
        \hline

        FM Covis   & 75.4                            & 87.6                             & 95.8 & 64.3 & 89.8 & 95.9 \\
        Pose Covis & 74.8                            & 86.9                             & 96.4 & 64.3 & 89.8 & 96.9 \\
        \hline
    \end{tabular}
    \mycaption{Ablation study of covisibility graph.}{
        Building the covisibility graph using frustum overlap performs similarly to utilizing feature matching.
        Accuracy at (0.25m, 2°), (0.5m, 5°), and (5m, 10°) thresholds.
    }
    \label{tab:covgraph_quality}
\end{table}

\begin{table}[t]
    \centering
    \begin{tabular}{c|rrr|rrr}
        \hline
        Supervision & \multicolumn{3}{c|}{ Dept. 1F Val} & \multicolumn{3}{c}{Dept. 4F Val}                             \\
        \hline
        Original    & 62.3                               & 82.2                             & 93.7 & 59.1 & 82.2 & 97.6 \\
        Adjusted    & 70.6                               & 86.6                             & 95.5 & 63.9 & 84.2 & 98.3 \\
        Depth       & 76.8                               & 88.1                             & 95.6 & 68.5 & 84.9 & 98.5 \\
        \hline
    \end{tabular}
    \mycaption{Ablation study of supervision methods.}{
        \textit{Original} refers to the original reprojection error supervision, \textit{Adjusted} refers to our depth-adjusted reprojection error supervision, and \textit{Depth} uses ground truth depth for supervision.
        Accuracy at (0.1m, 1°), (0.25m, 2°), and (1m, 5°).
        }
    \label{tab:supervision_accuracy}
\end{table}
\begin{figure}[t]
    \centering
    \includegraphics[width=\linewidth]{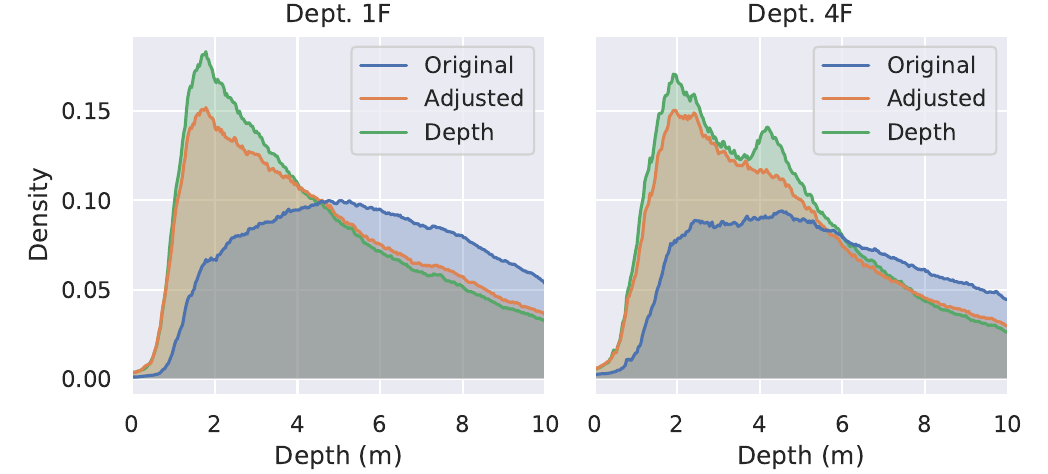}
    \mycaption{Ablation study of depth distribution after training with different supervision methods.}{ %
    Our depth-adjusted supervision matches the distribution of ground truth depth for supervision as compared to the original.
    }
    \label{fig:supervision_median_depth}
\end{figure}

\customparagraph{Supervision}
Our depth-adjusted supervision effectively mitigates the bias towards distant points and enhances the implicit triangulation of nearby points. As demonstrated in \myfigref{fig:supervision_median_depth}, depth-adjusted supervision significantly alters the depth distribution of predicted points, alleviating the previous ignorance of nearby points. This adjustment brings the distribution closer to that achieved with ground truth depth supervision, demonstrating a substantial reduction in the bias inherent in the original supervision approach.

In \mytabref{tab:supervision_accuracy}, we observe that depth-adjusted supervision also leads to notable improvements in localization accuracy, particularly under stricter thresholds, where accurate translation estimation relies heavily on near points.
Even without ground truth depth supervision, depth-adjusted supervision enables the model to achieve competitive performance.

\section{Conclusion}

In this work, we revisited scene coordinate regression (SCR) methods for robust visual localization in large-scale, complex environments. We analyzed the design principles of input encoding and training strategies, identifying several areas for enhancement. Our proposed R-SCoRe includes a covisibility graph-based global encoding learning and data augmentation strategy, a depth-adjusted reprojection loss to improve the implicit triangulation, and also other improvements including better architecture and local feature. Our contributions advance the state-of-the-art in SCR and demonstrate that SCR-based localization methods %
can achieve competitive performance in large-scale applications.
While operating at comparably very small map sizes, R-SCoRe trails the state-of-the-art FM-based localization methods only at the strictest error thresholds. Although out-of-distribution generalization remains challenging, and gaps persist in handling extreme cases, 
given the relatively small history of SCR, we are positive the accuracy gap can be closed completely in the near future.

{
    \small
    \bibliographystyle{ieeenat_fullname}
    \bibliography{main}
}

\clearpage
\setcounter{page}{1}
\maketitlesupplementary

\setcounter{section}{0}
\renewcommand{\thesection}{\Alph{section}}

In this supplementary, we first elaborate on the details in the implementation of R-SCoRe.
After that, we show additional results and interpret their meaning. 
Finally, we reflect on the current limitations of R-SCoRe and discuss future work we consider to improve the performance of localization with SCR further and close the gap to feature matching methods completely.

\section{Implementation Details}
\subsection{Local encodings}
\customparagraph{Pretrained feature extractor} For Dedode~\cite{edstedt2024dedode}, we select the top 5,000 keypoints per image using the Dedode-L detector and extract features using the Dedode-B descriptor. For LoFTR~\cite{sun2021loftr}, we utilize the CNN feature grid after layer 3, which is 8$\times$ smaller than the input image. We use the center of each grid cell as the keypoint.

\begin{figure}[b]
    \centering
    \includegraphics[width=\linewidth]{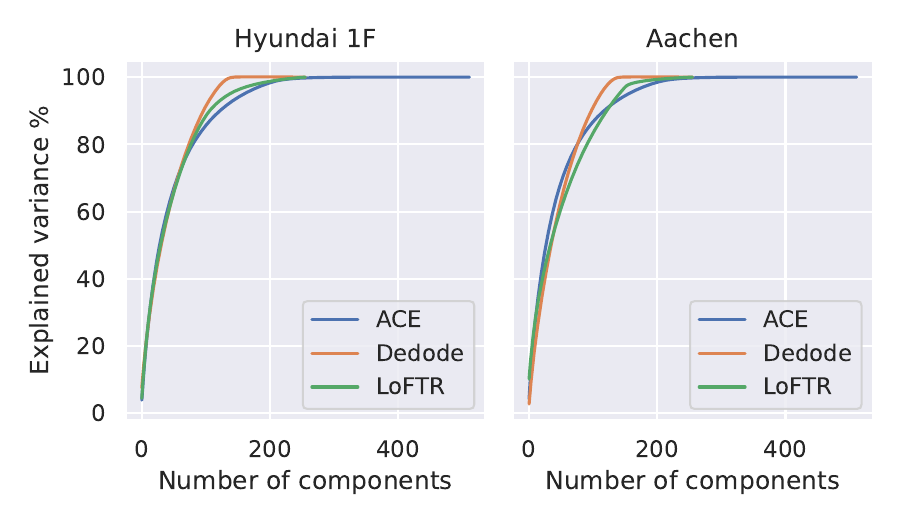}
    \mycaption{Local Encoding PCA.}{The ratio of variance explained by different numbers of PCA dimensions of local encodings.  Reducing the dimensionality to 128 dimensions usually preserves over 90\% of the variance.
    }
    \label{fig:local_pca}
\end{figure}

\customparagraph{Local encoding PCA} 
Before training, we run PCA on the local encodings to reduce their dimensionality to 128 entries. 
As shown in \myfigref{fig:local_pca}, reducing the feature dimensionality to 128 dimensions preserves over 90\% of the variance for different local encoders~\cite{brachmann2023ace,sun2021loftr,edstedt2024dedode} on various datasets~\cite{Sattler2012BMVC,Sattler2018CVPR,lee2021indoorlocdataset}.
To enable efficient computation of the PCA on the GPU, we extract approximately 10 million features via sampling from the training images.
In order to incorporate all available features, incremental PCA could be used instead. 
However, we found that sampling achieves similar performance. 

\customparagraph{Local encoding buffer} We allocate the training buffer with 32 million 128-dimensional features per GPU, across four GPUs, for a total of 128 million features in half-precision floating-point format. 

\customparagraph{Image data augmentation} 
Similar to previous works~\cite{glace2024cvpr,brachmann2023ace},
each image undergoes data augmentation with random resizing, rotation, and color jittering, before we extract local features. 
Random resizing adjusts the shorter edge, uniformly sampled between 320 and 720 pixels. Rotation is applied uniformly within the range of -15 to 15 degrees, while brightness and contrast are jittered with factors uniformly sampled from [0.9, 1.1]. %

\subsection{Global Encoding Learning with Node2Vec}
We use Node2Vec~\cite{node2vec-kdd2016} to learn node embeddings for the training images based on the covisibility graph of the scene. Node2Vec performs weighted random walks on the graph and learns embeddings with the Skip-gram~\cite{mikolov2013efficient} objective. The random walk is controlled by two parameters: the return parameter $p$, and the in-out parameter $q$. These parameters influence the random walk behavior: the probability of returning to the previous node is proportional to $\frac{1}{p}$, moving farther from the current node is proportional to $\frac{1}{q}$, and staying equidistant to the previous node is proportional to 1.

We use parameters favoring less exploration: $p = 0.25$ and $q = 4$. The embedding dimension is set to 256, aligning with the $R^2$Former~\cite{zhu2023r2former} feature dimension used in GLACE~\cite{glace2024cvpr} to enable a fair comparison in our evaluation.

\subsection{Covisibility Graph Construction}
We estimate covisibility directly from camera poses using a weighted frustum overlap, following~\cite{rau2020predictingoverlap, sarlin2022lamar}. For each image \( i \), we uniformly sample \( N_i \) pixels and unproject each with random depths within \([0, d_{\text{v}}]\), then check visibility \( V_k(i \to j) \) from viewing frustum image \( j \). The directed overlap score is computed as:
\begin{equation}
    O(i \to j) = \frac{\sum_{k=1}^{N_i} V_k(i \to j) \alpha_k(i,j)}{N_i},
\end{equation}
where \(\alpha_k(i,j)\) is the cosine similarity between ray directions. The covisibility graph is constructed by applying a threshold of 0.2 to the harmonic mean of \(O(i \to j)\) and \(O(j \to i)\). We use maximum viewing frustum depth $d_v = 8$ for indoor scenes and $d_v = 50$ for outdoor scenes.%

Recall that Table 6 of the main paper compares covisibility graph construction from frustum overlap 
to a more sophisticated version that performs feature matching. 
For the Aachen Day-Night~\cite{Sattler2012BMVC,Sattler2018CVPR}, we observe similar performance and, hence, 
prefer the simpler algorithm, based on frustum overlap. 
Here, we shed some light on how covisibility graph construction from feature matching is implemented. 
First, we perform feature matching between image pairs using SuperPoint~\cite{detone2018superpoint} 
and SuperGlue~\cite{sarlin2020superglue}, verified against ground truth poses. 
Second, we consider image pairs covisible that possess 100 or more matched keypoints.

\subsection{Network Architecture}
We adopt the MLP architecture and position decoder from GLACE~\cite{glace2024cvpr}, enhanced with an additional refinement module. The architecture employs \( n = 3 \) residual blocks for both the initial output and the refinement module, resulting in a total of six residual blocks. The width of the residual blocks is set to \( w = 768 \) for the Aachen~\cite{Sattler2012BMVC,Sattler2018CVPR} and Hyundai Department Store~\cite{lee2021indoorlocdataset} 4F datasets, and \( w = 1280 \) for the Hyundai Department Store~\cite{lee2021indoorlocdataset} B1 and 1F datasets. The hidden width in the residual block is expanded by a factor $m=2$.

\subsection{Training Details}
The training is conducted over 100,000 iterations using the AdamW~\cite{Loshchilov2019DecoupledWD} optimizer, with a weight decay set to 0.01. With 4 NVIDIA GeForce RTX 4090, the training takes approximately 4 hours for smaller networks with width $w = 768$ and up to 8 hours for larger networks with width $w = 1280$.
For additional acceleration and memory efficiency, our model is trained with mixed precision. Finally, the model weight and bias are saved in a half-precision format to reduce the model size. 
An exception are the training camera cluster centers, which are saved in single-precision.

\section{Additional Results}

\begin{figure*}[h]
    \centering
    \includegraphics[width=\linewidth]{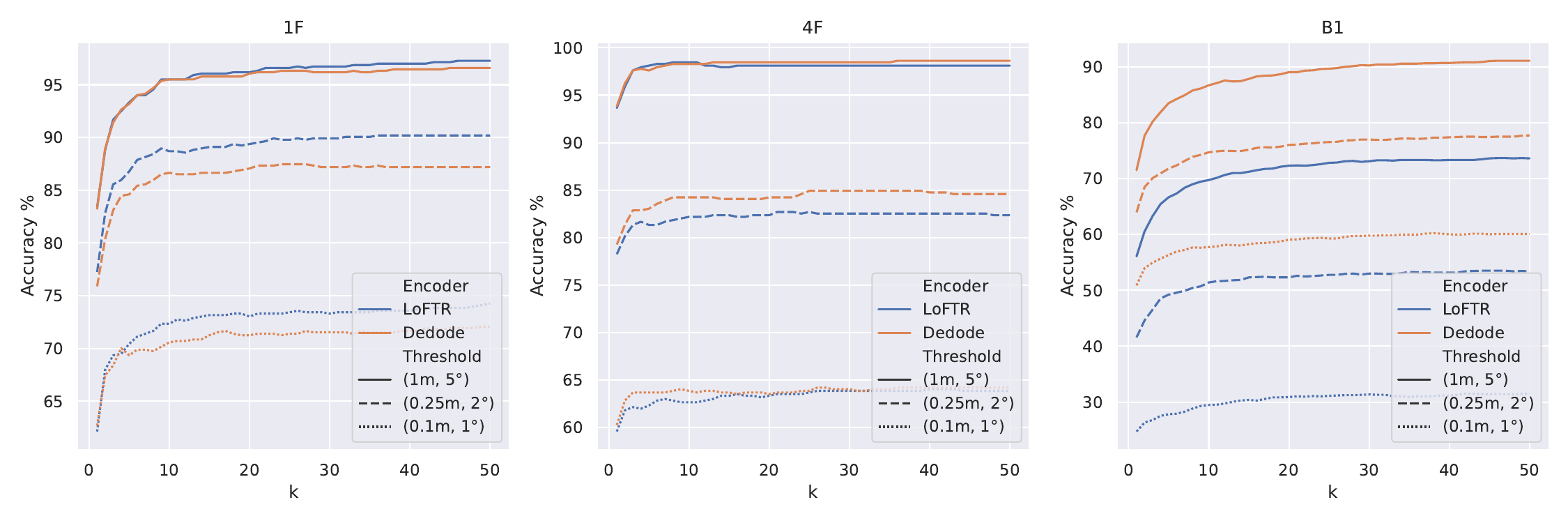}
    \mycaption{Comparison of localization accuracy with different number of global hypotheses.}{The accuracy at (0.1m, 1°), (0.25m, 2°), and (1m, 5°) thresholds with different numbers of global hypotheses is plotted. Increasing the number of hypotheses improves localization performance, though the performance gain typically plateaus when the number of hypotheses exceeds 10. }
    \label{fig:naver_knn_acc}
\end{figure*}

\begin{table}[h]
    \centering
    \begin{tabular}{cc|rrr}
        \hline
                        Encoding& Augmentation                                 & \multicolumn{3}{c}{Dept. 1F Val}               \\
        \hline
       $R^2$Former~\cite{zhu2023r2former}  & Gaussian  & 42.1 & 74.5 & 92.2 \\
$R^2$Former~\cite{zhu2023r2former}  & Covis  & 62.0 & 83.8 & 94.8 \\
Covis & Covis  & 72.3 & 88.7 & 95.5 \\
\hline
Covis & Gaussian & 59.1 & 78.9 & 90.5 \\
        \hline
    \end{tabular}
    \mycaption{Ablation study of global encodings.}{
        Accuracy at (0.1m, 1°), (0.25m, 2°), and (1m, 5°) thresholds. The isotropic Gaussian data augmentation can also work with our covisibility graph encoding directly, while the best performance is achieved by using our covisibility graph data augmentation.}
        
    \label{tab:ablation_global2}
\end{table}

\subsection{Hyundai Department Store Validation Results}
The results for the validation set of Hyundai Department Store~\cite{lee2021indoorlocdataset} are shown in \mytabref{tab:val_hyundai}. Note that Neumap~\cite{tang2023neumap} only provides their result on the validation set. 
In our main paper we evaluate on the official test set of~\cite{lee2021indoorlocdataset},
and, hence, \cite{tang2023neumap} is omitted from the evaluation there. 
The findings from the validation set are similar to the analysis we conduct in the main paper. 
While Neumap~\cite{tang2023neumap} delivers similar performance to R-SCoRe (using local encodings of Dedode~\cite{edstedt2024dedode}) on 1F and 4F, it significantly trails our method on B1. In addition, R-SCoRe  maintains about 6-8$\times$ smaller map sizes and its localization speed appears to be considerably faster than those of Neumap~\cite{tang2023neumap}.

\begin{table*}[h]
    \resizebox{\linewidth}{!}{

        \begin{tabular}{l|c|c|c}
            \hline
                                                                                 & Dept. 1F Validation                       & Dept. 4F Validation              & Dept. B1 Validation               \\
            \hline
            HLoc+D2-Net~\cite{sarlin2019coarse,DusmanuCVPR19D2NetDeepLocalFeatures}                       & (83.2 / 89.2 /94.5) /        398GB        & (72.1 / 85.3 / 98.5)   / 183GB   & (70.2/ 78.0 / 86.1) /  505GB      \\
            HLoc+R2D2~\cite{sarlin2019coarse,RevaudNIPS19R2D2ReliableRepeatableDetectorsDescriptors} & (85.8 / 89.9 / 94.4) /   166GB            & (72.6/ 84.6 / 98.3)   / 76GB     & (71.6/ 78.0  / 86.0) /     210GB  \\
            \hline
            PoseNet~\cite{kendall2015posenet}                           & (0.0 / 0.0 / 0.4) /                  41MB & (0.0 / 0.0 / 0.2)   / 41MB       & (0.0 / 0.0 / 0.0) /     41MB      \\
            \hline
            Neumap~\cite{tang2023neumap}                                & (75.5 / 88.2 / 95.8) / 726MB              & (70.4 / 85.4 / 99.0) / 431MB     & (46.0 /66.5 / 79.8) / 857MB       \\
            \hline
            ESAC ($\times50$)~\cite{brachmann2019esac}                                                        & (49.7  / 71.5  / 84.1) /1.4GB             & (45.2 /  69.9  / 85.1) / 1.4GB   & ( 5.4  / 9.1 / 14.2 ) /     1.4GB \\
            ACE ($\times50$)~\cite{brachmann2023ace}                             & (14.2 / 49.9 / 77.8) / 205MB              & (29.3 / 80.0 / 96.7) / 205MB     & (2.6 / 14.0 / 28.2) / 205MB       \\
            GLACE~\cite{glace2024cvpr}                                  & (4.9 / 24.4 / 53.5) / 42MB                & (24.5 / 57.5 / 85.4) / 42MB      & (1.0 / 4.5 / 13.8) / 42MB         \\
            \hline
            R-SCoRe (LoFTR$^*$~\cite{sun2021loftr})                                                              & (72.3 / 88.7 / 95.5) / 127MB              & (62.5 / 82.2 / 98.6) / 50MB      & (29.4 / 51.3 / 69.6) / 130MB      \\
            ~~+ Depth                                                          & (74.7 / 89.2 / 95.9) / 127MB              & (67.6 / 84.4 / 98.5) / 50MB      & (32.4 / 54.4 / 71.0) / 130MB      \\
           R-SCoRe (Dedode~\cite{edstedt2024dedode})                                                             & (70.6 / 86.6 / 95.5) / 127MB              & (63.9 / 84.2 / 98.3) / 50MB      & (57.7 / 74.7 / 86.7) / 130MB      \\
            ~~+ Depth                                                          & (77.1 / 88.6 / 95.6) / 127MB              & (68.5 / 84.9 / 98.5) / 50MB      & (59.5 / 75.6 / 86.8) / 130MB      \\
            \hline
        \end{tabular}
    }

    \mycaption{Hyundai Department Store Validation Set evaluation.}{The percentages of query images within three thresholds: (0.1m, 1°), (0.25m, 2°), and (1m, 5°) and the map size are reported.
    R-SCoRe achieves competitive accuracy with a small map size.
    $^*$We use LoFTR~\cite{sun2021loftr} outdoor, trained on MegaDepth~\cite{li2018megadepth}, instead of the indoor model trained on ScanNet~\cite{dai2017scannet} for the B1 scene with strong illumination change.
    }

    \label{tab:val_hyundai}
\end{table*}

\subsection{Additional Global Encoding Ablation}
As shown in \myfigref{fig:naver_knn_acc}, using multiple hypotheses can deliver a significant gain in performance.
In general, increasing the number of hypotheses improves the performance, although the gain diminishes when the number of hypotheses becomes larger than 10. 

In \mytabref{tab:ablation_global2}, we explore whether isotropic Gaussian data augmentation proposed in~\cite{glace2024cvpr} can also work with our covisibility graph encoding.
While we can indeed (\cf last row) improve the performance directly, our covisibility graph augmentation delivers better results for either encoding. 
For the experiment, we use the same standard deviation $\sigma=0.1$ for the noise as in GLACE~\cite{glace2024cvpr}.

\subsection{Network Architecture Ablation}
\begin{figure*}[t]
    \centering
    \includegraphics[width=\linewidth]{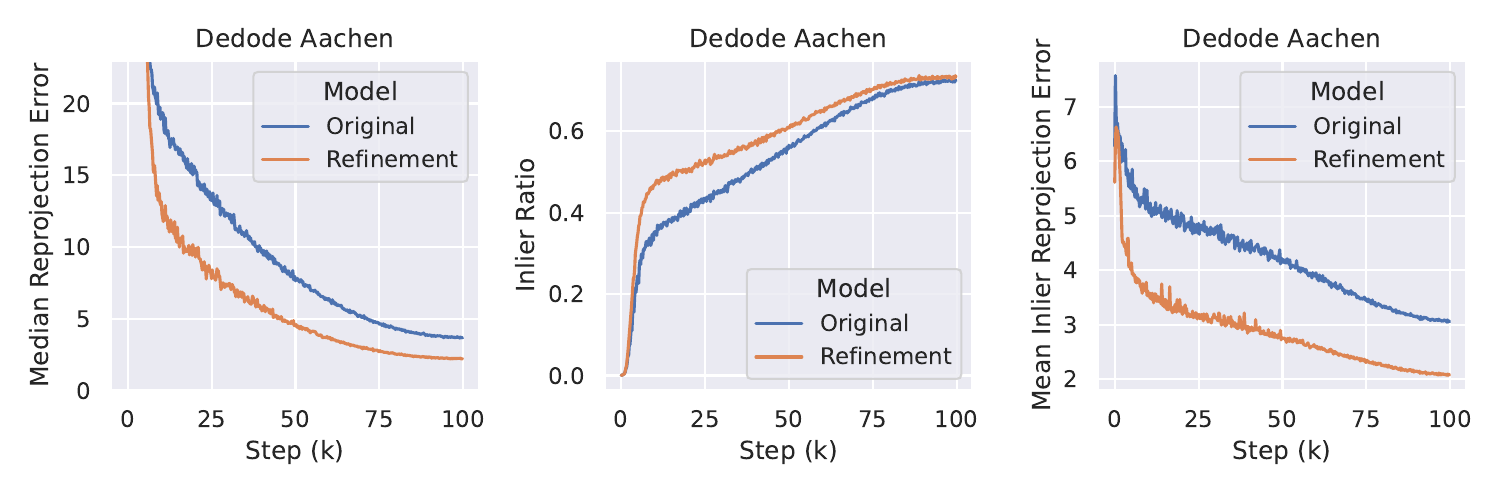}
    \mycaption{Ablation study of refinement module.}{We present the median reprojection error, the ratio of inlier training predictions with reprojection errors below 10 pixels, and the mean projection error of these inliers.}
    \label{fig:refinement_train}
\end{figure*}

Recall that our model predicts a coarse intermediate and a refined output. 
Without refinement, our network architecture becomes more similar to the standard SCR pipelines introduced in~\cite{brachmann2023ace,glace2024cvpr}.
To justify our design, we conduct an ablation study using the original network architecture without the refinement module. 
For a fair comparison, the baseline using the original architecture has the same total depth and width but directly outputs the final coordinate at the end without a coarse to fine refinement.
In training, our pipeline with the explicit refinement module 
achieves a lower median reprojection error and also reduces the training error more rapidly (\myfigref{fig:refinement_train}, left). 
Similarly, the ratio of inlier training predictions improves more quickly with explicit refinement, but after some time, both pipelines show a similar value (\myfigref{fig:refinement_train}, middle).
A closer look at the mean reprojection error (\myfigref{fig:refinement_train}, right) of these inliers shows a significant gap also at the end of training. 
We conjecture that our pipeline with the explicit refinement module can deliver more accurate predictions. 
Finally, as shown in \mytabref{tab:refinement_ablation}, the superior training performance also leads to improved localization accuracy of the pipeline with the explicit refinement module -- especially for stricter thresholds.
For this evaluation on Aachen Day-Night~\cite{Sattler2012BMVC,Sattler2018CVPR}, we employ covisibility graphs computed by frustum overlap. 

\begin{table}[t]
    \centering
    \begin{tabular}{l|ccc|ccc}
        \hline
                                        & \multicolumn{3}{c|}{Aachen Day} & \multicolumn{3}{c}{Aachen Night}                             \\
        \hline

        Original    &     65.5 & 82.9 & 95.3 & 51.0 & 78.6 & 96.9   \\
        Refinement      & 74.8 & 86.9 & 96.4 & 64.3 & 89.8 & 96.9  \\
        \hline
    \end{tabular}
    \mycaption{Ablation study of refinement module.}{Accuracy at  (0.25m, 2°), (0.5m, 5°), and (5m, 10°) thresholds are reported. The explicit refinement module improves the performance, especially for stricter thresholds.}
    \label{tab:refinement_ablation}
\end{table}

\section{Limitations and Future Work}
Throughout our evaluation, we show that R-SCoRe achieves competitive performance on recent large-scale benchmarks, while maintaining very small map sizes. 
Although we improve on recent SCR methods there still remains a gap -- compared to the state-of-the-art feature based methods -- in meeting the strictest pose quality thresholds. 
We conjecture that this limitation may stem from the network's inability to fully generalize and be invariant under extreme input variations, which makes the output coordinate not accurate enough. 
One potential direction for improvement is integrating our discriminative scene representation with generative models like NeRF~\cite{mildenhall2020nerf}. For instance, SCR could provide a robust initialization, which could then be refined by aligning with NeRF-based approaches~\cite{Chen_2024_CVPR_NeFes,yen2020inerf,zhou2024nerfmatch}.

Additionally, further reductions in map size could be explored by integrating techniques such as pruning~\cite{zhu2017prune},  low-rank approximation~\cite{rigamonti2013learning_lowrank}, and quantization~\cite{hubara2016binarizednetwork,polino2018quantization}, which all appear to be applicable to our pipeline in a straightforward manner.

\end{document}